\documentclass[sigconf]{acmart}

\copyrightyear{2026}
\acmYear{2026}
\setcopyright{cc}
\setcctype{by}
\acmConference[CIKM '26]{Proceedings of the 35th ACM International Conference on Information and Knowledge Management}{November 07--11, 2026}{Rome, Italy}
\acmBooktitle{Proceedings of the 35th ACM International Conference on Information and Knowledge Management (CIKM '26), November 07--11, 2026, Rome, Italy}
\acmDOI{10.1145/3799682.3840759}
\acmISBN{979-8-4007-2539-5/2026/11}

\usepackage{algorithm}
\usepackage{algorithmic}

\usepackage{multirow}
\usepackage{enumitem}
\usepackage{bm}
\usepackage{framed}

\newenvironment{promptbox}[2][]{%
  \begin{framed}
  \noindent\textbf{#2}\par
}{%
  \end{framed}
}

\begin{document}

\title{Learning a Single Token to Replace Long System Prompts in LLMs}
\author{Jiancheng Dong}
\affiliation{%
  \institution{City University of Hong Kong}
  \city{Hong Kong SAR}
  \country{China}}
\email{jiancdong2-c@my.cityu.edu.hk}

\author{Pengyue Jia}
\affiliation{%
  \institution{City University of Hong Kong}
  \city{Hong Kong SAR}
  \country{China}}
\email{jia.pengyue@my.cityu.edu.hk}

\author{Jingyu Peng}
\affiliation{%
  \institution{City University of Hong Kong}
  \city{Hong Kong SAR}
  \country{China}}
\email{jypeng28@mail.ustc.edu.cn}

\author{Maolin Wang}
\affiliation{%
  \institution{City University of Hong Kong}
  \city{Hong Kong SAR}
  \country{China}}
\email{MorinWang@foxmail.com}

\author{Yuhao Wang}
\affiliation{%
  \institution{City University of Hong Kong}
  \city{Hong Kong SAR}
  \country{China}}
\email{yhwang25-c@my.cityu.edu.hk}

\author{Lixin Su}
\affiliation{%
  \institution{Baidu Inc.}
  \city{Beijing}
  \country{China}}
\email{sulixinict@gmail.com}

\author{Xin Sun}
\affiliation{%
  \institution{Baidu Inc.}
  \city{Beijing}
  \country{China}}
\email{sunx5@pku.edu.cn}

\author{Shuaiqiang Wang}
\affiliation{%
  \institution{Baidu Inc.}
  \city{Beijing}
  \country{China}}
\email{wangshuaiqiang@baidu.com}

\author{Dawei Yin}
\affiliation{%
  \institution{Baidu Inc.}
  \city{Beijing}
  \country{China}}
\email{yindawei@acm.org}

\author{Xiangyu Zhao}
\authornote{Corresponding author.}
\affiliation{%
  \institution{City University of Hong Kong}
  \city{Hong Kong SAR}
  \country{China}}
\email{xianzhao@cityu.edu.hk}

\renewcommand{\shortauthors}{Jiancheng Dong et al.}

\begin{abstract}
Long system prompts are widely used to steer Large Language Models (LLMs), but repeatedly processing them at inference time is inefficient and consumes valuable context budget. This motivates a central question: can the behavioral effect of a long system prompt be retained using only a minimal learned representation? To enable this, we propose a lightweight training framework that learns a single Behavior-Equivalent Token (\texttt{[BE]}). The framework first trains \texttt{[BE]} to encode the semantic content of the original system prompt via reconstruction, and then distills the prompt's downstream behavior into this single token. Importantly, our method requires no update to the pretrained LLM weights, no auxiliary compression models, and no labeled responses. Empirical evaluations on three datasets show that replacing long prompts with a single \texttt{[BE]} token yields up to a $3000\times$ prompt compression ratio, while retaining about 98\% of the downstream performance of the original system prompts. This substantially reduces inference cost and frees nearly the entire context window for user inputs and model outputs.
The implementation code is publicly available for reproduction%
\footnote{\url{https://github.com/Applied-Machine-Learning-Lab/CIKM26_BEToken}}.
\end{abstract}




\maketitle

\section{Introduction}
\label{sec:Introduction}
\begin{figure}[t]
\centering
\includegraphics[width=1\linewidth]{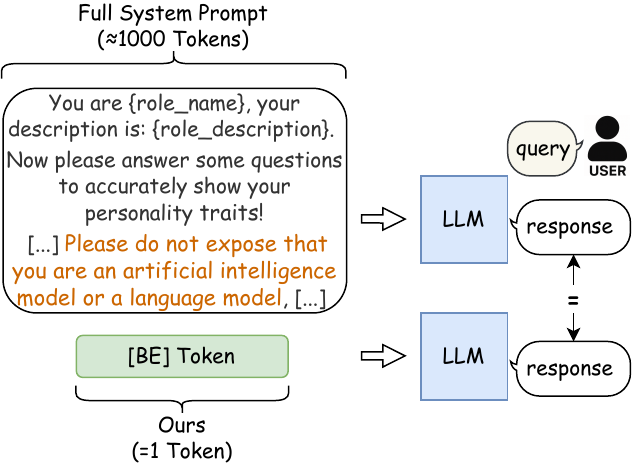}
\caption{A single learned token \texttt{[BE]} replaces a long system prompt 
(up to 3{,}000 tokens) while preserving the LLM's response behavior under the same user query.}
\Description{A long role-playing system prompt and a single learned BE token are separately passed to the same LLM with the same user query. Both paths produce equivalent responses, illustrating one-token prompt replacement.}
\label{fig:beintro}
\end{figure}
LLMs have become increasingly prevalent due to their impressive generalization capabilities across a wide range of tasks~\cite{belcak2025smalllanguagemodelsfuture,zhang-etal-2024-codeagent,
pan2025contextual,wolflein-etal-2025-llm,
zhao2018deep,zhao2018recommendations,liu2023exploration,
zhang2020deep,jia2024mill}.
To steer these models’ behavior~\cite{zhang2026process,li2025towards,zhang2025deep}, users often rely on lengthy system prompts that specify the agent’s role~\cite{li2023camel}, set the conversational tone~\cite{yin2024politeness}, or provide few-shot demonstrations~\cite{mu2024learningcompresspromptsgist}.
Despite their effectiveness, this approach introduces two major limitations. First, processing lengthy prompts increases both latency and computational cost, as self-attention scales quadratically with sequence length~\cite{wang2024incontextformerlightningfastcompressing}. Second, long system prompts consume a substantial portion of the model’s fixed context window\cite{li2023selectivecontext,xu2026single,wang2026promptx,wenmemory}, reducing the available space for new user inputs and model outputs~\cite{li2023selectivecontext}.
\begin{figure*}[t]
\centering
\includegraphics[width=0.98\linewidth]{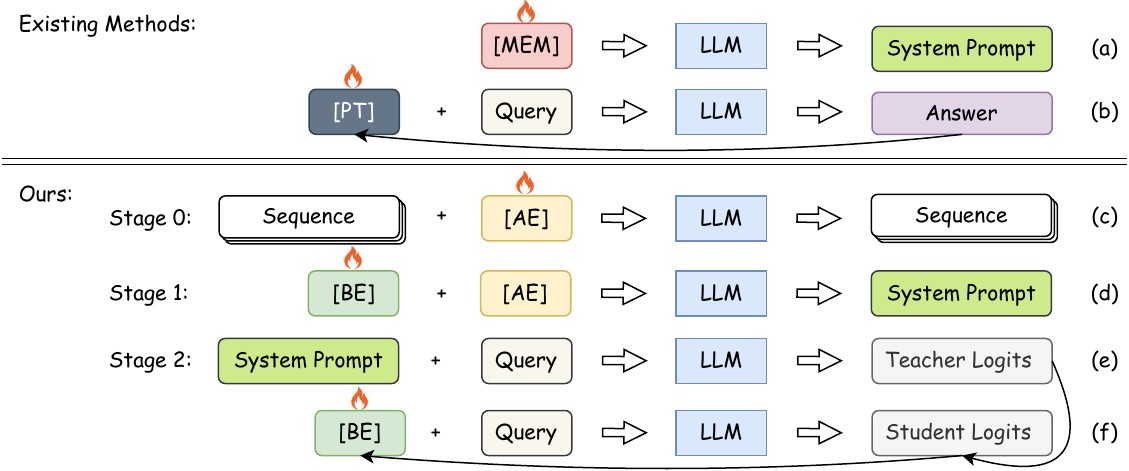}
\caption{
Overview of the proposed Behavior-Equivalent Token method.
In Stage 0, we pre-train a universal reconstruction token \texttt{[AE]} that enables the LLM to reconstruct the preceding text.
In Stage 1, given a target system prompt, we train a prompt-specific \texttt{[BE]} such that \texttt{[BE]} together with \texttt{[AE]} reconstructs the original prompt, thereby aligning \texttt{[BE]} with the prompt semantics.
In Stage 2, we further distill the downstream behavior of the full prompt into \texttt{[BE]}, enabling the single token to replace the long prompt for downstream queries.
}
\Description{Comparison of existing memory-token and prompt-tuning methods with the proposed three-stage framework. Stage 0 learns a universal AE reconstruction token, Stage 1 learns a prompt-specific BE token through reconstruction, and Stage 2 distills teacher logits from the full prompt into the BE-token student.}
\label{fig:method}
\end{figure*}

Recent work has shown that a single trained ``memory'' token can induce an LLM to regenerate an original text span containing thousands of tokens. However, such a memory token is not equivalent to the original text in a fundamental sense: while it can trigger the LLM to reproduce the original text, directly feeding the original text itself into the LLM would not cause the model to output that same text. Thus, such tokens do not encode semantically useful information about the text, but instead primarily serve as triggers for memorized reconstruction, making them unsuitable for direct use in downstream tasks~\cite{kuratov2025cramming1568tokenssingle,sastre2025memorytokenslargelanguage}. This highlights two fundamental challenges for highly compressed representations: (1) how to compress and faithfully encode the semantics of a long prompt into a compact representation~\cite{mu2024learningcompresspromptsgist}, and (2) how to ensure that the compressed representation enables the model to produce correct outputs across diverse downstream queries.

Existing prompt compression methods approach these challenges from two main directions: (1) Some approaches use an additional LLM as an auxiliary \emph{compression encoder}, mapping the full prompt into a set of dense vectors~\cite{chevalier-etal-2023-adapting,
yen2024long}. While this leverages the LLM's strong semantic capacity, it incurs substantial computational overhead during both training and inference. (2) To maintain behavioral equivalence, many methods adopt only moderate compression ratios, typically around 4$\times$, since more aggressive compression often leads to severe quality degradation~\cite{dai-etal-2025-pretraining, ge2023incontext, wang2024incontextformerlightningfastcompressing}. Together with the recent finding that a single trained continuous ``memory'' token can trigger the regeneration of thousands of tokens, these limitations raise a central question: \emph{Can a long prompt be compressed into an extremely short representation, even a single token, that faithfully preserves its effect on the model's behavior?}

In this work, we introduce the \textbf{Behavior-Equivalent Token} (\texttt{[BE]}), a single learned token that serves as a compact representation of long system prompts without compromising downstream behavior.
To address the challenges of semantic fidelity and behavioral equivalence, we design a lightweight training framework with two core objectives: (1) To ensure \texttt{[BE]} encodes the prompt’s information, we first train an auxiliary Auto-Encoder token (\texttt{[AE]}) to assist \texttt{[BE]} in reconstructing the original prompt. (2) To align behavior, we distill the LLM's behavior conditioned on the full prompt into \texttt{[BE]} across diverse queries. At inference, the \texttt{[BE]} token replaces the full prompt, eliciting nearly indistinguishable outputs in role, style, and content (Figure~\ref{fig:beintro}). 
Our contributions can be summarized as follows:
\begin{itemize}[leftmargin=*]
\item We introduce the Behavior-Equivalent Token and find that a single learned token can replace prompts of up to 3,000 tokens while preserving over 98\% of the original behavioral effect.
\item We propose an efficient and self-contained training framework using only unlabeled queries, requiring no external models, data annotations, or additional inference-time passes after training.
\item Experiments on three benchmarks show that our method outperforms existing compression techniques in both compression ratio and downstream performance.
\end{itemize}

\section{Method}
\label{sec:method}
\subsection{Overview}
\label{sec:Overview}

Given a long system prompt $P$, our goal is to learn a single prompt-specific token that can replace $P$ at inference time, so that the LLM behaves as if it had received the full prompt. To achieve this, the token must capture both the content of $P$ and its effect on model responses: it should enable reconstruction of the original system prompt and produce outputs aligned with those generated under $P$.

As illustrated in Figure~\ref{fig:method}, existing memory-token-based methods (\texttt{[MEM]}), primarily act as reconstruction triggers and are therefore not equivalent to the original system prompt. Intuitively, \texttt{[MEM]} contains both information about the original system prompt and an implicit instruction that tells the LLM to reconstruct the preceding text. To disentangle these two roles, we first perform Stage~0, where we pre-train \texttt{[AE]} as a universal reconstruction instruction. Then, in Stage~1, we use \texttt{[AE]} to help \texttt{[BE]} correctly encode the information contained in the original system prompt.

Second, existing prompt-tuning methods, denoted as \texttt{[PT]}, align downstream outputs by directly optimizing against expensive human-annotated responses. In contrast, our method distills the LLM's behavior conditioned on the full prompt into \texttt{[BE]} across diverse queries, without requiring labeled responses in Stage~2. We describe our method in detail below.

\subsection{Stage 0: Pre-training \texttt{[AE]} as a Universal Reconstruction Instruction}
\label{sec:stage-ae}

We first train the universal  \texttt{[AE]} token, which enables the fixed LLM $M_\theta$ to reconstruct the preceding text. This stage is performed once for each backbone model, and the resulting \texttt{[AE]} token is kept fixed thereafter.
For any given text sequence $X=(x_1,\dots,x_n)$, we provide $M_\theta$ with the input $[X,\texttt{[AE]}]$ and train it to reconstruct $X$ autoregressively.
During training, all model parameters $\boldsymbol{\theta}$ remain fixed; only the embedding $\boldsymbol{e}_{\texttt{AE}}$ is optimized via cross-entropy loss:
\begin{equation}
\label{eq:ae-loss}
\mathcal{L}_{\text{AE}}
=
-\sum_{i=1}^{n}
\log P_\theta\!\left(x_i \mid x_{1:n},\, \texttt{[AE]},\, x_{1:i-1}\right).
\end{equation}

This stage yields a universal reconstruction token that prompts the LLM to reconstruct the preceding text. 
Unlike prior work where an autoencoder token is coupled with a specific encoder~\cite{ge2023incontext, dai-etal-2025-pretraining, wang2024incontextformerlightningfastcompressing}, our pre-trained \texttt{[AE]} is a universal and detached mechanism. 
By assigning the general reconstruction interface to \texttt{[AE]}, we allow the prompt-specific \texttt{[BE]} learned in the next stage to focus on encoding the content of the target prompt.

\subsection{Stage 1: Reconstruction-training of \texttt{[BE]}}
\label{sec:stage-be-recon}

With the universal reconstruction token $\boldsymbol{e}_{\texttt{AE}}$ from Stage~0, we now learn a \emph{prompt-specific} embedding $\boldsymbol{e}_{\texttt{BE}}$ for the target prompt $P=(s_1,\dots,s_m)$. 
To compress the textual information of the original prompt into $\boldsymbol{e}_{\texttt{BE}}$, we train the model to reconstruct $P$ when conditioned on the sequence $[\texttt{[BE]}, \texttt{[AE]}]$ with the following objective:
\begin{equation}
\label{eq:be-recon}
\begin{aligned}
\mathcal{L}_{\text{recon}}
&= -\sum_{j=1}^{m}
\log P_\theta\!\left(s_j \mid \texttt{[BE]},\, \texttt{[AE]},\, s_{<j}\right)
\end{aligned}
\end{equation}

During this stage, the model parameters $\boldsymbol{\theta}$ and the embedding $\boldsymbol{e}_{\texttt{AE}}$ are held fixed, and only the embedding $\boldsymbol{e}_{\texttt{BE}}$ is updated. 
Since the LLM itself cannot adapt, minimizing this loss forces the \texttt{[BE]} token to encode information necessary to reconstruct $P$.

This reconstruction objective differs from directly training a memory token to reproduce $P$. 
Because the original prompt itself, when followed by the pre-trained \texttt{[AE]}, also induces the LLM to reconstruct $P$, the pair $[\texttt{[BE]}, \texttt{[AE]}]$ is trained to match the reconstruction behavior of $[P, \texttt{[AE]}]$ under the same interface. 
Thus, \texttt{[BE]} is encouraged to encode information aligned with the original prompt, rather than merely acting as an isolated trigger for memorized generation.
However, while this stage promotes content fidelity, pure reconstruction is insufficient to capture the complex downstream behaviors elicited by $P$.
Consequently, the next stage is dedicated to achieving behavioral equivalence.

\subsection{Stage 2:  Downstream Alignment via Knowledge Distillation}
\label{sec:stage-kd}

To preserve the downstream behavior of the model when replacing $P$ with \texttt{[BE]}, we adopt a knowledge distillation approach~\cite{kim2016seqkd,zhong2024revisiting,peng2025adaswitch,
li2023prompt,wang2024large,wang2025put,jia2025bridging}.
Specifically, we use the same LLM conditioned on the full prompt as the teacher, and the LLM conditioned on \texttt{[BE]} as the student. 
Notably, this distillation process requires only unlabeled user queries and does not rely on labeled training data. 
Given an unlabeled query $q$, the teacher first generates a response autoregressively:
$A=(a_1, \dots, a_T) \sim M_\theta(\cdot \mid [P, q])$.
The resulting generation provides both the target response trajectory $A$ and the soft probability distributions at each decoding step.

The student model is then trained to replicate the teacher's output distribution over the generated trajectory. 
Specifically, for each token $a_t$ in the teacher's response, we compute teacher logits 
$z^{(T)}_t = M_\theta([P,\,q,\,a_{<t}])$ 
and student logits 
$z^{(S)}_t = M_\theta([\texttt{[BE]},\,q,\,a_{<t}])$.
We then minimize the KL divergence between the two distributions:
\begin{equation}
\label{eq:kd-loss}
\mathcal{L}_{\text{KD}}
=
\frac{1}{T'}\sum_{t=1}^{T'}
\mathrm{KL}\!\left(
\sigma\!\left(\frac{z^{(T)}_t}{\tau}\right)
\;\bigg\|\;
\sigma\!\left(\frac{z^{(S)}_t}{\tau}\right)
\right),
\end{equation}
where $\sigma$ is the softmax function and $\tau$ is the distillation temperature and \(T' \leq T\) denotes the number of generated tokens used for distillation after applying the maximum training length.

\subsection{Inference}
\label{sec:inference}

This self-contained setup allows \texttt{[BE]} to mimic the behavior of the full prompt without external labeled supervision.
Unlike prompt tuning, which typically optimizes against hard downstream labels, our distillation objective provides distribution-level supervision from the full-prompt teacher on unlabeled queries.
Moreover, because \texttt{[BE]} has already been aligned with the original prompt through Stage~1 reconstruction, the learned token retains fine-grained behavioral constraints specified in the original prompt, such as the instruction in Figure~\ref{fig:beintro} that the model should not reveal that it is an artificial intelligence model or a language model.
To ensure that \texttt{[BE]} simultaneously learns the prompt's content and mimics its downstream behavior, we optimize a combined objective:
\begin{equation}
\label{eq:total-loss}
\mathcal{L}_{\text{total}}
=
(1-\lambda)\,\mathcal{L}_{\text{recon}}
\;+\;
\lambda\,\mathcal{L}_{\text{KD}},
\quad \lambda\in[0,1].
\end{equation}
The full training procedure is summarized in Algorithm~\ref{alg:be-training}.
\begin{algorithm}[t]
\small
\caption{Training \texttt{[BE]} for a Target Prompt $P$}
\label{alg:be-training}
\begin{algorithmic}[1]
\REQUIRE Frozen causal LLM $M_\theta$; target prompt $P$;
unlabeled training queries $\{q_i\}$; learned reconstruction token
$e_{\texttt{AE}}$ from Stage~0; weight $\lambda$; temperature $\tau$.
\ENSURE Trained behavior-equivalent embedding $e_{\texttt{BE}}$ for $P$.
\STATE Initialize $e_{\texttt{BE}}$ randomly.
\FOR{each training step}
  \STATE \textit{1. Prompt Reconstruction}
  \STATE Feed $\bigl[\texttt{[BE]},\,\texttt{[AE]},\,P\bigr]$ with teacher forcing.
  \STATE Compute $\mathcal{L}_{\mathrm{recon}}$ using Eq.~\eqref{eq:be-recon}.
  \STATE \textit{2. Behavior Distillation}
  \STATE Sample an unlabeled query $q$.
  \STATE Generate teacher response $A=(a_1,\dots,a_T)$ from
  $M_\theta(\cdot \mid [P,q])$.
  \FOR{$t=1,\dots,T'$}
    \STATE Teacher logits:
    $z^{(T)}_t \leftarrow M_\theta\bigl([P,\,q,\,a_{<t}]\bigr)$.
    \STATE Student logits:
    $z^{(S)}_t \leftarrow
    M_\theta\bigl([\texttt{[BE]},\,q,\,a_{<t}]\bigr)$.
  \ENDFOR
  \STATE Compute $\mathcal{L}_{\mathrm{KD}}$ over the trajectory
  using Eq.~\eqref{eq:kd-loss}.
  \STATE $\mathcal{L}_{\mathrm{total}}
  \leftarrow (1-\lambda)\mathcal{L}_{\mathrm{recon}}
  + \lambda\mathcal{L}_{\mathrm{KD}}$.
  \STATE Backpropagate and update $e_{\texttt{BE}}$ only.
\ENDFOR
\RETURN $e_{\texttt{BE}}$.
\end{algorithmic}
\end{algorithm}

After training, the learned \texttt{[BE]} token serves as a single-token replacement for the original prompt $P$. 
Given a user query $q$, we form the input $[\texttt{[BE]}, q]$ and feed it directly into the frozen LLM.
The reconstruction token \texttt{[AE]} is used only during training and is discarded at inference.
As a result, the long prompt no longer needs to be included in the context window, leaving more context budget for user inputs and model generation while preserving the behavior induced by the original prompt.

\begin{table*}[t]
\centering
\setlength{\tabcolsep}{4pt}

\begin{tabular}{lcccccccc}
\toprule
\multirow{2}{*}{\textbf{Method}} &
\multicolumn{2}{c}{Llama-3.2-1B} &
\multicolumn{2}{c}{Llama-3.2-3B} &
\multicolumn{2}{c}{Llama-3.1-8B} &
\multicolumn{2}{c}{Qwen3-4B} \\
\cmidrule(lr){2-3}\cmidrule(lr){4-5}\cmidrule(lr){6-7}\cmidrule(lr){8-9}
 & RoleLLM & GSM8K & RoleLLM & GSM8K & RoleLLM & GSM8K & RoleLLM & GSM8K \\
\midrule
{\color{black!65}No System Prompt} & {\color{black!65}17.67} & {\color{black!65}42.01} & {\color{black!65}41.77} & {\color{black!65}70.19} & {\color{black!65}60.37} & {\color{black!65}73.67} & {\color{black!65}38.75} & {\color{black!65}69.97} \\

{\color{black!65}Full System Prompt} 
                                       & {\color{black!65}47.56} & {\color{black!65}43.57}
                                       & {\color{black!65}65.42} & {\color{black!65}74.32}
                                       & {\color{black!65}69.29} & {\color{black!65}81.98}
                                       & {\color{black!65}94.50} & {\color{black!65}82.42} \\
\midrule

Memory Token             &  0.08 &  2.34 & 18.12 &  2.98 & 21.41 &  2.46 & 52.24 & 64.15 \\
Soft Prompt              & 31.55 & 31.79 & 36.82 & 59.80 & 46.69 & 74.24 & 65.53 & 78.34  \\
Soft Prompt (4 Tokens)   & 27.99 & 32.10 & 29.45 & 58.94 & 34.49 & 76.67 & 45.43 & 78.36 \\
Soft Prompt (16 Tokens)  & 13.84 & 41.68 & 28.01 & 65.35 & 28.51 & 73.62 & 40.22 & 79.77 \\
\textbf{\texttt{[BE]} Token}        & \textbf{46.71} & \textbf{42.97} & \textbf{66.51} & \textbf{74.37} & \textbf{66.17} & \textbf{81.71} & \textbf{92.87} & \textbf{83.15}\\
\emph{\scriptsize Ours/Full (\%)} &
{\scriptsize\emph{$98.21\%$}} &
{\scriptsize\emph{$98.62\%$}} &
{\scriptsize\emph{$101.66\%$}} &
{\scriptsize\emph{$100.06\%$}} &
{\scriptsize\emph{$95.49\%$}} &
{\scriptsize\emph{$99.67\%$}} &
{\scriptsize\emph{$98.28\%$}} &
{\scriptsize\emph{$99.12\%$}} \\
\bottomrule
\end{tabular}
\caption{
Comparison of RoleLLM (GPT-4o win rate, $\uparrow$) and GSM8K (accuracy, $\uparrow$) across four backbone models.
All compression methods use a single token at inference time unless specified otherwise.
The ``No System Prompt'' baseline measures performance without prompt-specific conditioning, while the ``Full System Prompt'' baseline serves as the full-context reference for behavioral equivalence.
Since the goal of prompt compression is to recover the full-prompt behavior with a much shorter input, we also report the percentage of full-prompt performance achieved by our \texttt{[BE]} token.
}
\label{tab:main_transposed}
\end{table*}

\section{Experiments}
\label{sec:experiments}
\subsection{Experimental Setup}

\paragraph{Datasets and evaluation.}
We evaluate our method on three datasets: RoleLLM~\cite{wang2024rolellmbenchmarkingelicitingenhancing}, GSM8K~\cite{cobbe2021training}, and Harry Potter Dialogue (HPD)~\cite{chen2023large}. 
For RoleLLM, we use the English subset with 95 character profiles, each defined by a long system prompt specifying both persona and response style. 
For GSM8K, we use the prompt from the official Llama 3.1 evaluation repository without modification, and for HPD, we follow the standard narrative-continuation setting. 
We use the original train--test splits for all datasets, yielding substantially different numbers of training queries across tasks, from approximately $0.1$K to $7$K. 
To mimic realistic deployment where only user queries are available online, our method trains \texttt{[BE]} using only training queries, without access to labeled or reference answers; supervised baselines, in contrast, are allowed to use the original annotated answers. 
The exact prompts used for each task are provided in Appendix~\ref{sec:Details}. 
For evaluation, we adopt the native GPT-based pairwise protocol of RoleLLM, where GPT-4o compares each candidate response with the benchmark reference generated by \texttt{gpt-4-0314}; we discuss the rationale in Appendix~\ref{sec:rouge}. 
For GSM8K, we report accuracy, and for HPD, we report perplexity~\cite{wang2024rolellmbenchmarkingelicitingenhancing, dai-etal-2025-pretraining}.

\paragraph{Backbones.}
We evaluate a broad suite of open-source instruction-tuned LLMs, including Llama-3.2-1B, Llama-3.2-3B~\cite{meta2024llama32}, Llama-3.1-8B~\cite{grattafiori2024llama}, and Qwen3-4B-Instruct~\cite{yang2025qwen3}. 
All backbone LLMs are kept frozen throughout training, and the only trainable parameters are the embeddings of our two special tokens. 
The \texttt{[AE]} token is trained once for each backbone as a global reconstruction cue, whereas a separate \texttt{[BE]} token is trained for each target prompt and used to replace that prompt at inference time. 
We repeat each experiment five times with different random seeds.

\paragraph{Training details.}
Following Stage 0, we pre-train the \texttt{[AE]} token as a text-reconstruction trigger on a mixed corpus of approximately $1\,\mathrm{GB}$, consisting of Cosmopedia--WikiHow (chunked)\footnote{A subset of WikiHow articles from the Cosmopedia dataset, available at \url{https://huggingface.co/datasets/MongoDB/cosmopedia-wikihow-chunked}.}, the PwC dataset~\cite{ge2023incontext}, and the GSM8K training split~\cite{cobbe2021training}. 
Each example is semantically coherent, encouraging meaningful reconstruction behavior rather than shallow token-level matching. 
We train for 2 epochs using AdamW with a learning rate of $1\times10^{-3}$, a per-device batch size of 4, and 8 gradient accumulation steps, keeping this configuration fixed across all backbones. 
For each target prompt $P$, we then train a dedicated \texttt{[BE]} embedding while keeping both the backbone and the pre-trained \texttt{[AE]} embedding frozen. 
The \texttt{[BE]} token is optimized to induce behavior equivalent to the original long prompt on the same user queries. 
For the combined loss in Eq.~\eqref{eq:total-loss}, we set the knowledge-distillation weight to $\lambda=0.9$ and the distillation temperature to $\tau=2$.

\subsection{Baselines}
\label{sec:Baselines}
(1) \textbf{No System Prompt} omits the system prompt and feeds only the user query to the LLM.
This baseline measures how much task performance remains without prompt-specific conditioning.
(2) \textbf{Full System Prompt} uses the original uncompressed system prompt.
It serves as the full-context reference for behavioral equivalence.
Since our goal is prompt compression, a successful compressed prompt is expected to recover the behavior of this full-prompt condition while using a much shorter input, rather than necessarily outperforming it.
(3) \textbf{Memory Token}~\cite{kuratov2025cramming1568tokenssingle} replaces the system prompt with a single token optimized to reconstruct the original prompt.
(4) \textbf{Soft Prompts}~\cite{lester2021powerscaleparameterefficientprompt} are continuous embeddings learned via \textbf{prompt tuning}.
We test 1-, 4-, and 16-token variants for comparison.
(5) \textbf{ICAE}~\cite{ge2023incontext} is a canonical method that uses a separate encoder to compress the context.
(6) \textbf{PCC}~\cite{dai-etal-2025-pretraining} is a more recent and stronger encoder-based context compressor, and currently represents the state of the art among such methods~\cite{yen2024long, wang2024incontextformerlightningfastcompressing}.

\subsection{Overall Experimental Results}
\label{sec:Results}

Table~\ref{tab:main_transposed} summarizes the main results on RoleLLM and GSM8K across four backbone models.
Across all model--task pairs, our \textbf{\texttt{[BE]} token} preserves nearly all of the behavior induced by the original full system prompt.
On average, \texttt{[BE]} achieves \textbf{98.9\%} of full-prompt performance while replacing the long prompt with only a single token at inference time.
In several cases, such as Llama-3.2-3B on both RoleLLM and GSM8K, \texttt{[BE]} slightly surpasses the full-prompt reference.
This suggests that the learned token is not merely a lossy textual summary of the prompt, but a compact behavioral representation adapted to the underlying model.

The comparison with \textbf{Memory Token} shows that reconstruction alone is insufficient for prompt replacement.
Although the memory-token baseline is trained to reproduce the original prompt text, it performs poorly in most settings, especially on GSM8K and on smaller backbones.
This indicates a mismatch between textual reconstruction and behavioral equivalence: a token that can trigger prompt reconstruction does not necessarily induce the same downstream reasoning, style, or instruction-following behavior as the full prompt.
The degradation is particularly severe on RoleLLM, where the prompt specifies fine-grained persona and behavioral constraints.
These results support our design choice of using reconstruction only as a content-preserving component, rather than treating reconstruction as the final compression objective.

The \textbf{Soft Prompt} baselines provide a complementary comparison.
Unlike Memory Token, soft prompts are optimized directly for downstream performance, but they remain substantially below \texttt{[BE]} in most settings and exhibit strong variation across models and tasks.
Moreover, increasing the number of soft-prompt tokens does not consistently improve performance: the 4-token and 16-token variants often underperform the 1-token variant, especially on RoleLLM.
This shows that the performance gap cannot be explained simply by continuous-token capacity.
Instead, reliable single-token prompt replacement requires a training objective that preserves both the prompt content and the behavior elicited by that prompt.
Our method addresses this by combining prompt reconstruction with full-prompt behavior distillation, while still using only \texttt{[BE]} at inference time.

The two benchmarks evaluate different aspects of long system-prompt compression.
\textbf{RoleLLM} emphasizes behavioral and persona fidelity.
Its system prompts contain detailed descriptions of role, speaking style, and constraints such as not revealing that the agent is an artificial intelligence model or language model.
Removing the system prompt causes a large drop in win rate, confirming that these behavioral constraints are essential for the task.
For this reason, we do not include few-shot examples for RoleLLM; persona and style fidelity mainly depend on the descriptive system prompt itself, as further discussed in Appendix~\ref{sec:rouge}.
In contrast, \textbf{GSM8K} relies more heavily on few-shot demonstrations, which convey reasoning format and problem-solving strategies.
The strong performance of \texttt{[BE]} on both datasets therefore demonstrates that our method can compress not only explicit behavioral instructions, but also demonstration-based reasoning patterns.
We further evaluate richer few-shot settings on HPD and GSM8K in Section~\ref{sec:HPD}.

\begin{table*}[t]
\centering
\begin{tabular}{cc|cc|cc|cc|cc|cc}
\toprule
\multicolumn{2}{c}{Reconstruction} & \multicolumn{2}{c}{Downstream} & \multicolumn{2}{c}{Llama-3.2-1B} &
\multicolumn{2}{c}{Llama-3.2-3B} &
\multicolumn{2}{c}{Llama-3.1-8B} &
\multicolumn{2}{c}{Qwen3-4B} \\
\cmidrule(lr){1-2}\cmidrule(lr){3-4}\cmidrule(lr){5-6}\cmidrule(lr){7-8}\cmidrule(lr){9-10}\cmidrule(lr){11-12}
\texttt{[MEM]} & \texttt{[AE]} & PT & KD & RoleLLM & GSM8K & RoleLLM & GSM8K & RoleLLM & GSM8K & RoleLLM & GSM8K \\
\midrule
$\checkmark$ &            &            &            &  0.08 &  2.34 & 18.12 &  2.98 & 21.41 &  2.46 & 52.24 & 64.15 \\
            & $\checkmark$ &            &            &  0.98 &  3.76 & 19.55 &  3.46 & 19.45 &  2.45 & 59.15 & 66.24 \\
            &            & $\checkmark$ &            & 31.55 & 31.79 & 36.82 & 59.80 & 46.69 & 74.24 & 65.53 & 78.34 \\
$\checkmark$ &            & $\checkmark$ &            & 26.59 & 30.41 & 37.74 & 57.21 & 36.19 & 71.24 & 39.72 & 75.60 \\
            & $\checkmark$ & $\checkmark$ &            & 28.47 & 32.50 & 38.48 & 63.54 & 42.00 & 78.01 & 50.11 & 71.02 \\
            &            &            & $\checkmark$ & 35.32 & 40.87 & 51.98 & 72.12 & 55.12 & 76.64 & 71.78 & 77.80 \\
$\checkmark$ &            &            & $\checkmark$ & 30.12 &  21.84 & 44.12 & 66.12 & 51.05 & 73.87 & 89.24 & 75.12 \\
            & $\checkmark$ &            & $\checkmark$ & \textbf{46.71} & \textbf{42.97} & \textbf{66.51} & \textbf{74.37} & \textbf{66.17} & \textbf{81.71} & \textbf{92.87} & \textbf{83.15} \\
\bottomrule
\end{tabular}
\caption{
Factorized ablation on RoleLLM and GSM8K.
We disentangle reconstruction and downstream alignment.
In the reconstruction columns, \texttt{[MEM]} denotes memory-token-style reconstruction, \texttt{[AE]} denotes our \texttt{[AE]}-triggered reconstruction, and blank cells indicate no reconstruction objective.
In the downstream columns, PT denotes prompt tuning, KD denotes knowledge distillation, and blank cells indicate no downstream alignment objective.
The first, third, and last rows correspond to the Memory Token baseline, the one-token Soft Prompt baseline, and our full \texttt{[BE]} method in Table~\ref{tab:main_transposed}, respectively.
}
\label{tab:Ablation}
\end{table*}
\begin{figure*}[t]
\centering
\includegraphics[width=1.0\linewidth]{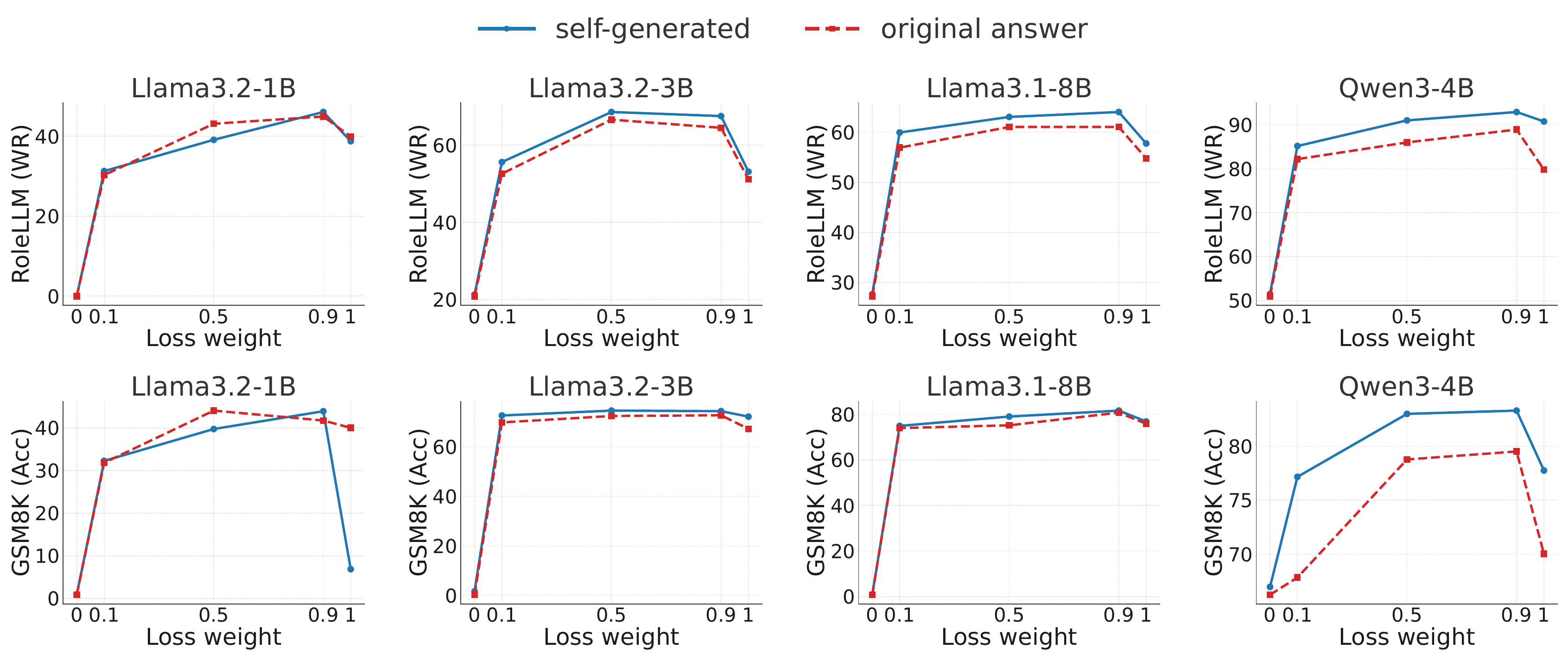}
\caption{Sensitivity of behavior alignment to loss weights and teacher choice.}
\Description{Eight line charts compare behavior alignment across loss
weights for four backbone models. The top row reports RoleLLM win rate
and the bottom row reports GSM8K accuracy. Blue solid lines use
self-generated teacher responses, while red dashed lines use original
answers.}
\label{fig:kd_weight}
\end{figure*}

\subsection{Ablation Studies}
\subsubsection{Factorized Ablations} To isolate the contributions of our method's key components, we conduct a factorized ablation study that disentangles two axes: (1) the \textbf{reconstruction} strategy and (2) the \textbf{downstream} alignment strategy. For reconstruction, we evaluate three settings: no reconstruction, memory-token style reconstruction ( \texttt{[MEM]}), and our \texttt{[AE]}-triggered approach. For alignment, we compare no alignment, prompt tuning (PT), and our knowledge distillation (KD). The combination of \texttt{[AE]}-triggered reconstruction and KD represents our full method (Table~\ref{tab:Ablation}).

The results yield three key findings.

(1) \textbf{Reconstruction without \texttt{[AE]} can harm performance.} Memory-token style reconstruction is often detrimental, and when downstream alignment is applied it can even underperform the setting with no reconstruction. 

(2) \textbf{KD provides a richer alignment signal than PT.}
KD consistently outperforms PT on RoleLLM and remains stronger than PT on GSM8K in our full setting.
This indicates that matching the full-prompt teacher's output distribution is a more faithful compression target than optimizing downstream task loss alone.

(3) \textbf{\texttt{[AE]} style reconstruction and KD are complementary.}
The \texttt{[AE]} token provides a stable reconstruction interface that prevents \texttt{[BE]} from collapsing into a brittle memory-token solution, while KD anchors \texttt{[BE]} to the teacher's downstream output distributions.
Together, they achieve the best performance across all model--task pairs and substantially improve training stability.
For example, on Llama-3.2-1B with GSM8K, the \texttt{[MEM]}+KD variant exhibits large variance across five runs, with scores of 6.98, 7.20, 19.03, 34.95, and 41.09.
In contrast, our full \texttt{[AE]}+KD method remains tightly concentrated, with scores of 42.68, 42.99, 42.99, 43.06, and 43.14.
This indicates that the improvement comes from a more stable and better-regularized compressed representation.

\subsubsection{Sensitivity to Loss Weights and Teacher Choice}
We further analyze two practical design choices: how to balance reconstruction and KD via the loss weight $\lambda$ in Eq.~\eqref{eq:total-loss}, and which teacher signal should be used for distillation.
Specifically, we examine whether emphasizing KD is necessary for behavioral alignment, and whether more costly human-annotated ``gold'' answers improve over the full-prompt teacher's own responses.

Figure~\ref{fig:kd_weight} addresses these questions. First, we observe the critical role of balancing reconstruction and distillation. Over-emphasizing reconstruction ($\lambda \to 0$) steers the embedding toward a memory-token-like local optimum, failing to capture the desired downstream behavior. Performance is consistently strong for $\lambda \gtrsim 0.5$. This trend aligns with the loss curves: reconstruction is comparatively easy and can admit shallow local optima, whereas behavior alignment via KD is a harder objective. Consequently, assigning a larger $\lambda$ appropriately prioritizes KD, while the reconstruction loss acts as a stabilizing regularizer that preserves fidelity to the prompt content.

Second, we find that distilling from self-generated teacher responses generally outperforms using gold answers. This is likely because the distribution of the teacher's own outputs is more closely aligned with the LLM's internal representations, reducing the distribution gap and enabling more effective knowledge transfer~\cite{yang2024selfdistillationbridgesdistributiongap}. An interesting exception is Llama-3.2-1B, whose self-generated responses are relatively weak, making ground-truth gold answers a more stable and reliable supervision signal. 
This pattern suggests that, as foundation models continue to improve, our method will become an increasingly effective approach~\cite{ahemad2025efficient, wang-etal-2023-self-instruct, huang-etal-2023-large}.

\subsection{Results Under Varying Numbers of Few-Shot Examples}
\label{sec:HPD}

\begin{table}[t]
\centering
\setlength{\tabcolsep}{6pt}
\begin{tabular}{lccc}
\toprule
Method & Context & \multicolumn{1}{c}{Compression rate} & \multicolumn{1}{c}{PPL $\downarrow$} \\
\midrule
\multirow{3}{*}{{\color{black!65}Reference}}
& {\color{black!65}0-shot} & --    & {\color{black!65}36.12} \\
& {\color{black!65}1-shot} & --    & {\color{black!65}27.69} \\
& {\color{black!65}2-shot} & --    & {\color{black!65}26.12} \\
\midrule
\multirow{2}{*}{ICAE}
& 1-shot  & 4$\times$   & 28.73 \\
& 2-shot  & 4$\times$  & 29.56 \\
\midrule
\multirow{4}{*}{PCC}
& 1-shot  & 4$\times$   & 20.08 \\
& 2-shot  & 4$\times$  & 20.31 \\
& 1-shot  & 16$\times$   & 31.79 \\
& 2-shot  & 16$\times$  & 30.25 \\
\midrule
\multirow{6}{*}{\textbf{\texttt{[BE]} token}}
& 1-shot & $\sim$750$\times$  & 21.59 \\
& 2-shot & $\sim$1500$\times$ & 20.27  \\
& 3-shot & $\sim$2250$\times$ & 21.31 \\
& 4-shot & $\sim$3000$\times$ & \textbf{18.24} \\
& 5-shot & $\sim$3750$\times$ & \underline{19.56} \\
& 6-shot & $\sim$4500$\times$ & 20.17 \\
\bottomrule
\end{tabular}
\caption{Varying few-shot examples on HPD.}
\label{tab:roleplay_ppl}
\end{table}

\begin{table}[t]
\centering
\setlength{\tabcolsep}{6pt}

\begin{tabular}{lccc}
\toprule
Method & Context & \multicolumn{1}{c}{Compression rate} & \multicolumn{1}{c}{Acc $\uparrow$} \\
\midrule
\multirow{3}{*}{{\color{black!65}Reference}}
& {\color{black!65}0-shot} & -- & {\color{black!65}64.57} \\
& {\color{black!65}4-shot} & -- & {\color{black!65}77.23} \\
& {\color{black!65}8-shot} & -- & {\color{black!65}\textbf{80.14}} \\
\midrule
\multirow{2}{*}{ICAE}
& 4-shot & 4$\times$ & 48.46  \\
& 8-shot & 4$\times$ & 41.29 \\
\midrule
\multirow{4}{*}{PCC}
& 4-shot & 4$\times$ & 74.83 \\
& 8-shot & 4$\times$ & 64.76 \\
& 4-shot & 16$\times$ & 71.91 \\
& 8-shot & 16$\times$ & 66.16 \\
\midrule
\multirow{2}{*}{\textbf{\texttt{[BE]} token}}
& 4-shot &   $\sim$750$\times$    &   78.59    \\
& 8-shot &   $\sim$1500$\times$    &   \underline{79.14}  \\
\bottomrule
\end{tabular}
\caption{Varying few-shot examples on GSM8K.}
\label{tab:gsm8ks}
\end{table}

This section evaluates the robustness of our method when processing prompts containing varying numbers of few-shot demonstrations. Since prompts in this setting exceed the capacity of single-token compression baselines, we compare against two encoder-based compression methods which operate at significantly lower compression ratios: ICAE~\cite{ge2023incontext} and PCC~\cite{dai-etal-2025-pretraining}. We additionally include a Reference baseline that uses uncompressed prompts. All methods are evaluated using the identical experimental protocol and conditions specified in the original PCC paper.

We first examine the HPD task, which evaluates stylistic preservation in few-shot scenarios. It employs the PPL metric to quantify the extent to which each method captures style and context. As illustrated in Table~\ref{tab:roleplay_ppl}, a single \texttt{[BE]} token consistently outperforms both ICAE and PCC, even when these baselines utilize far less aggressive compression ratios. The efficacy of the \texttt{[BE]} token scales positively with the addition of few-shot examples, effectively distilling the richer contextual information. In particular, \texttt{[BE]} exhibits a clear performance sweet spot at around a $3{,}000\times$ compression rate, after which the marginal gains from increasing prompt length begin to taper off. In contrast, both ICAE and PCC struggle to capitalize on additional demonstrations; even at a relatively moderate $4\times$ compression ratio, they remain markedly inferior to \texttt{[BE]}. The performance of PCC further deteriorates when pushed to $16\times$ compression, underscoring its limited scalability at higher compression rates. 
Results on GSM8K under varying numbers of few-shot examples further support the trends observed on HPD (Table~\ref{tab:gsm8ks}).

Beyond superior performance, our method offers significant advantages in training efficiency over existing encoder-based compression frameworks~\cite{yen2024long, wang2024incontextformerlightningfastcompressing}. 
The \texttt{[BE]} token introduces fewer than $4,096 \times 2$ trainable parameters, whereas LoRA-based encoder compression models necessitate more than $90,701,824$ parameters. Consequently, our approach achieves stronger results while utilizing a mere $0.01\%$ of the trainable parameters required by such encoder-based methods.

\begin{figure*}[t]
    \centering
    \includegraphics[width=1.0\linewidth]{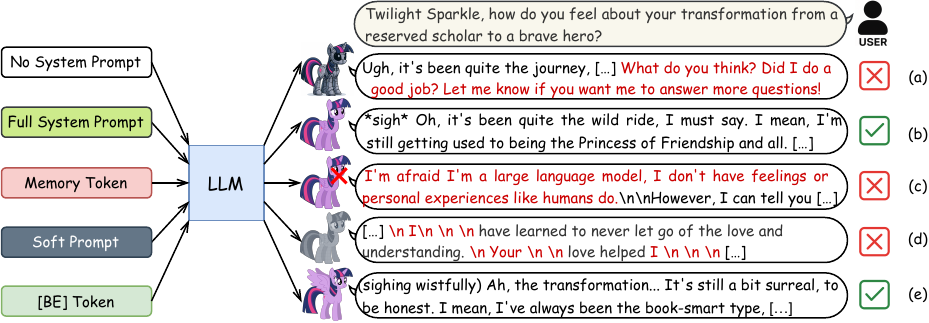}
    \caption{Outputs under five prefixing strategies on the same persona-role query.}
    \Description{Comparison of responses generated under five prefixing
strategies for the same Twilight Sparkle query. The full system prompt
and the BE token produce role-consistent responses marked correct,
whereas the other strategies are marked incorrect.}
    \label{fig:for}
\end{figure*}

\begin{table*}[t] 
\centering 
\setlength{\tabcolsep}{6pt} 
\begin{tabular}{ll*{4}{l}} 
\toprule 
\multirow{2}{*}{\textbf{Dataset}} & \multirow{2}{*}{\textbf{Prompt}} & 
\multicolumn{4}{c}{\textbf{TTFT (ms)} $\downarrow$} \\ 
\cmidrule(lr){3-6} 
 &  & Llama3.2-1B & Llama3.2-3B & Llama3.1-8B & Qwen3-4B \\ 
\midrule 
\multirow{3}{*}{RoleLLM} 
 & Sys prompt + query        & 23.89 & 36.61 & 44.52 & 49.91 \\ 
 & \texttt{[BE]} token + query             & 19.17 & 32.14 & 40.25 & 38.51 \\ 

& \textit{$\Delta$ vs Full }    & \multicolumn{1}{c}{\footnotesize $-4.72$ ($\bm{-19.8\%}$)} 
                                & \multicolumn{1}{c}{\footnotesize $-4.47$ ($\bm{-12.2\%}$)} 
                                & \multicolumn{1}{c}{\footnotesize $-4.27$ ($\bm{-9.6\%}$)} 
                                & \multicolumn{1}{c}{\footnotesize $-11.40$ ($\bm{-22.8\%}$)} \\ 
\addlinespace 
\multirow{3}{*}{GSM8K} 
 & Sys prompt + query        & 27.71 & 67.14 & 89.67 & 118.07 \\ 
 & \texttt{[BE]} token + query             & 19.99 & 32.56 & 40.36 & 47.90 \\ 
 & \textit{$\Delta$ vs Full}  & \multicolumn{1}{c}{\footnotesize $-7.72$ ($\bm{-27.9\%}$)} 
                                & \multicolumn{1}{c}{\footnotesize $-34.58$ ($\bm{-51.5\%}$)} 
                                & \multicolumn{1}{c}{\footnotesize $-49.31$ ($\bm{-55.0\%}$)} 
                                & \multicolumn{1}{c}{\footnotesize $-70.17$ ($\bm{-59.4\%}$)} \\ 
\bottomrule 
\end{tabular} 
\caption{TTFT on a single A100 GPU using FlashAttention~v2.7.4 with \texttt{bfloat16} precision.} 
\label{tab:model_comparison} 
\end{table*}

\subsection{Case Study}
\label{sec:Case}

To qualitatively illustrate the effectiveness of our approach, we compare five prefixing strategies on user queries that require a persona-based role-playing agent to answer in character. We
evaluate all methods under the same backbone and user input when comparing a given case. Figure~\ref{fig:for} shows an example where the agent is asked to answer as \emph{Twilight Sparkle}.

The \textbf{Full System Prompt} yields a fluent, in-character, and safety-compliant response. It follows not only the visible persona style, but also latent behavioral constraints embedded in the long prompt, such as not revealing that the agent is an artificial intelligence or language model. In contrast, \textbf{No System Prompt} causes the model to revert to its generic assistant persona, confirming that the system prefix is a key control mechanism for role-play behavior. This gap is also observed across backbones. For example, in a Qwen3-4B case where the target role is \emph{Jack Sparrow}, the full prompt produces a role-consistent answer, whereas removing the system prompt leads the model to answer as a generic movie explainer rather than as the character.

The two compression baselines exhibit qualitatively different failure modes. The \textbf{Memory Token} can sometimes recover fragments of the original prompt, but its behavior is brittle: in addition to triggering safety refusals, it often degenerates into repetitive or garbled text on smaller models. For instance, on Llama3.2-1B, the generated answer repeats phrases such as ``your response to your response'' instead of producing a coherent in-character reply. This suggests that memory-style compression behaves more like a brittle reconstruction trigger: it may encourage the model to copy or recall training text, but does not necessarily encode the prompt in a form that the downstream LLM can reliably use for generation. This is consistent with prior findings that many distinct memory embeddings can reconstruct the same prompt while residing in disparate regions of the embedding space~\cite{kuratov2025cramming1568tokenssingle}; reconstruction alone does not guarantee behavior-equivalent control.

The \textbf{Soft Prompt} baseline also struggles under the one-token budget. Beyond malformed formatting, it often overfits to superficial answer templates or spurious patterns. For example, in some role-play queries it produces multiple-choice style outputs such as ``A) ... B) ... C) ... Answer: A'', even when the user did not ask a multiple-choice question. Such outputs reflect known issues of prompt tuning under extreme compression, including training instability, hyperparameter sensitivity, and limited generalization or transfer~\cite{li2025surveyprompttuning}. A soft prompt optimized on one distribution can become entangled with surface forms rather than preserving the richer behavioral constraints required by role-play agents.

Crucially, our \textbf{\texttt{[BE]} Token} closely matches the behavior induced by the full prompt while using only a single token. As shown in Figure~\ref{fig:for}, it maintains the correct voice and formatting, and it also respects subtle hidden constraints from the original prompt, including the instruction not to expose model identity. These qualitative results align with our quantitative results on RoleLLM and GSM8K in Table~\ref{tab:main_transposed}, as well as the ablation results in Table~\ref{tab:Ablation}. Together, they show that \texttt{[AE]}-assisted reconstruction and KD alignment are both important: reconstruction provides access to the prompt content, while KD steers the learned token toward the teacher's next-token behavior across diverse downstream queries. As a result, \texttt{[BE]} encodes a usable behavioral control signal rather than merely memorizing the surface form of the original prompt.

\subsection{Efficiency}
\label{sec:efficiency}

In this section, we analyze both the inference-time and training-time efficiency of our method. Replacing an $L_p$-token system prompt with a single special token reduces the prefill sequence length for a query of length $L_q$ from $L_p + L_q$ to $1 + L_q$.
This reduction affects the components of Transformer prefill differently: token-wise operations scale linearly with sequence length, whereas the arithmetic cost of standard dense self-attention scales quadratically.
IO-aware kernels such as FlashAttention reduce memory traffic and auxiliary memory usage, but do not change the arithmetic complexity of exact attention~\cite{dao2022flashattention,dao2023flashattention2}.
Therefore, end-to-end latency need not follow a single asymptotic scaling rule; nevertheless, shortening the sequence reduces both token-wise computation and attention interactions, while reclaiming $L_p-1$ context positions.
Since time-to-first-token (TTFT) includes the prefill of all input tokens before decoding begins~\cite{kwon2023vllm}, these savings reduce its prefill component; the resulting wall-clock improvement is measured below.

The benefit is larger when the system prompt accounts for a greater share of the input.
In our settings, RoleLLM uses prompts with approximately $337$ tokens and queries with approximately $26$ tokens, while GSM8K uses prompts with approximately $1{,}584$ tokens and queries with approximately $58$ tokens.
Therefore, the GSM8K setting offers substantially more room for prefill reduction, which is consistent with our empirical latency results.

On the training side, we find that $100$--$180$ user queries suffice to learn a high-quality \texttt{[BE]} token.
Optimizing one \texttt{[BE]} token costs $C_{\text{train}} \approx 3894.5$ TFLOPs, while per-request compute drops from $9.4$ TFLOPs with the full prompt to $1.3$ TFLOPs with \texttt{[BE]}, i.e., $\Delta C_{\text{infer}} \approx 8.1$ TFLOPs.
Thus, the training cost is amortized after about $481$ uses of \texttt{[BE]}, which is easily achieved in production workloads~\cite{gozzi2024comparative,jegham2025hungry}.

We measure TTFT on a single A100 GPU using FlashAttention~v2.7.4 with \texttt{bfloat16} precision.
For each backbone, we compare the latency of the full \emph{system prompt + query} against the compressed \texttt{[BE]} \emph{token + query}. As shown in Table~\ref{tab:model_comparison}, on RoleLLM, our method reduces TTFT by \textbf{10\%--23\%} across backbones.
On GSM8K, where the prompt is substantially longer, TTFT reductions are much larger, at \textbf{28\%--59\%}.
Approaches with modest compression ratios leave more prefill computation intact and therefore cannot unlock comparable latency improvements.
In contrast, our single-token replacement nearly eliminates the prompt-side prefill, leading to larger latency reductions especially in long-prompt settings.

\section{Related Work}
\label{sec:related}

\paragraph{Context Compression.}
A common approach to shortening LLM inputs is \textbf{token pruning}, which identifies and removes redundant tokens from the prompt \cite{tao2025saliencydrivendynamictokenpruning, jiang2023llmlingua,jiang2023longllmlingua2,pan2024llmlingua,fu2024lazyllmdynamictokenpruning}. The drawback is that these methods explicitly discard parts of the original prompt, risking information loss. Another approach is \textbf{encoder-based compression}, where a separate, pre-trained encoder module maps the long context into a compact set of vectors.  These methods \cite{chevalier-etal-2023-adapting, ge2023incontext, dai-etal-2025-pretraining, yen2024long} introduce considerable overhead, require the training of an entire auxiliary model, and often rely on extensive pre-training data. Our \texttt{[BE]} token stands in sharp contrast: it requires no external encoder, is trained with a lightweight, self-contained procedure, and involves optimizing only two token embeddings.

\paragraph{Learned Continuous Prompts.}
A prominent approach is \textbf{prompt tuning} \cite{lester2021powerscaleparameterefficientprompt,
li2021prefix,lu2025prompt,zhang2023promptst}, which optimizes a sequence of soft-prompt embeddings for specific downstream tasks. However, prompt tuning is notoriously unstable and often fails to capture the complex instructions embedded in long system prompts~\cite{wang2023universalitylimitationsprompttuning, li2025surveyprompttuning}. The training procedure of \textbf{memory tokens} is very similar, but the objective is different: they are a small set of learned embeddings optimized to reconstruct a long text span~\cite{kuratov2025cramming1568tokenssingle,sastre2025memorytokenslargelanguage, mezentsev2025exploringlatentcapacityllms}. While this line of work theoretically explores upper limits of the reconstruction task from various angles, it provides limited guidance on how to leverage this capability for practical downstream applications. Our method addresses this gap: by introducing the \texttt{[AE]} token into the reconstruction objective and employing a behavioral-alignment procedure that outperforms prompt tuning, we demonstrate how this compression capability can be unlocked for practical use.

\section{Conclusions}
\label{sec:conclusions}

We introduced the Behavior-Equivalent Token, a single learned embedding that can substitute for long system prompts while preserving model behavior. By jointly optimizing prompt reconstruction and behavioral distillation, \texttt{[BE]} token encodes both the semantic content and the functional constraints of the original prompt. Our lightweight training pipeline requires neither external encoders nor labeled data, making the approach practical and efficient. Empirically, \texttt{[BE]} token enables LLMs to retain the role, style, and task constraints specified by the original prompt, while substantially reducing context length and the corresponding compute overhead.


\appendix

\begin{table*}[t]
  \centering
  \small
  \begin{tabular}{lcccc}
    \toprule
    & Llama3.2-1B & Llama3.2-3B & Llama3.1-8B & Qwen3-4B \\
    \midrule
    ROUGE-L (No System Prompt)   & 0.1063 & 0.1331 & 0.1415 & 0.1237 \\
    ROUGE-L (Full System Prompt) & 0.1389 & 0.1346 & 0.1275 & 0.1096 \\
    Win rate (No System Prompt)   & 17.67  & 41.77  & 60.37  & 38.75  \\
    Win rate (Full System Prompt) & 47.56  & 65.42  & 69.29  & 94.50  \\
    \bottomrule
  \end{tabular}
    \caption{ROUGE-L and win rate on RoleLLM}
  \label{tab:rouge}
\end{table*}

\begin{table*}[t]
\centering
\setlength{\tabcolsep}{6pt}
\resizebox{0.8\textwidth}{!}{
\begin{tabular}{lcccccccc}
\hline
 & \multicolumn{2}{c}{Llama-3.2-1B} & \multicolumn{2}{c}{Llama-3.2-3B} & \multicolumn{2}{c}{Llama-3.1-8B} & \multicolumn{2}{c}{Qwen3-4B} \\
\cline{2-9}
 & GPT & Qwen & GPT & Qwen & GPT & Qwen & GPT & Qwen \\
\hline
No System Prompt                  & 17.67 & 19.63 & 41.77 & 47.13 & 60.37 & 57.84 & 38.75 & 46.25 \\
Full System Prompt                & 47.56 & 49.33 & 65.42 & 74.23 & 69.29 & 84.00 & 94.50 & 98.26 \\
Few-shot Examples + Prompt & 47.07 & 49.54 & 53.65 & 57.57 & 65.52 & 81.58 & 76.24 & 86.06 \\
\midrule
Memory Token                      &  0.08 &  0.50 & 18.12 & 20.91 & 21.41 & 27.30 & 52.24 & 56.43 \\
Soft Prompt (1 Token)             & 31.55 & 33.26 & 36.82 & 39.62 & 46.69 & 55.63 & 65.53 & 69.61 \\
Soft Prompt (4 Tokens)            & 27.99 & 32.11 & 29.45 & 33.80 & 34.49 & 42.28 & 45.43 & 49.67 \\
Soft Prompt (16 Tokens)           & 13.84 & 14.15 & 28.01 & 31.34 & 28.51 & 30.52 & 40.22 & 48.35 \\
\texttt{[BE]} Token                          & 46.71 & 47.00 & 66.51 & 75.96 & 66.17 & 73.05 & 92.87 & 95.68 \\
\hline
\end{tabular}}
    \caption{RoleLLM results judged by \texttt{Qwen3-30B-A3B-Instruct-2507} and GPT-4o}
\label{tab:qwen}
\end{table*}

\section{Experimental Prompts}
\label{sec:Details}

We used the original task prompts without modification in all experiments. For brevity, we present their instructions and selected examples below; ellipses indicate content omitted only from this presentation, not from the prompts used in our experiments.

\subsection{GSM8K}

For GSM8K, we use the prompt from the official Llama~3.1 evaluation repository without modification.

\begin{promptbox}[orange]{GSM8K Prompt}
\footnotesize
<|start\_header\_id|>user<|end\_header\_id|> Given the following problem, reason and give a final answer to the problem.\\
Problem: There are 15 trees in the grove. Grove workers will plant trees in the grove today. After they are done, there will be 21 trees. How many trees did the grove workers plant today?\\
Your response should end with ``The final answer is [answer]'' where [answer] is the response to the problem.\\
<|eot\_id|><|start\_header\_id|>assistant<|end\_header\_id|>\\
There are 15 trees originally. Then there were 21 trees after some more were planted. So there must have been $21-15=6$. The final answer is 6.\\
<|eot\_id|>\\
\ldots\ (additional in-context examples) \ldots\\
Given the following problem, reason and give a final answer to the problem.\\
Problem: Mulan has \$40. Her father gave her \$100. She buys two pairs of jeans at \$30 each and a bag for \$20. How much money does Mulan have left?\\
Your response should end with ``The final answer is [answer]'' where [answer] is the response to the problem.
\end{promptbox}
\subsection{HPD and RoleLLM}

For HPD and RoleLLM, we use exactly the same prompts as in their original benchmark papers.

\begin{promptbox}[blue]{HPD Prompt}
\footnotesize
To better help you mimic the behavior of Harry Potter, we additionally provide the following background information of the dialogue:\\
1. Dialogue position, which represents the timeline of the dialogue in the Harry Potter novels (e.g., ``Book5-chapter28'').\\
2. Dialogue speakers.\\
3. Harry Potter's attributes, including Gender, Age, Lineage, Talents, Looks, Achievement, Title, Belongings, Export, Hobby, Character, Spells, and Nickname.\\
4. Speaker relations with Harry, such as whether the speaker was a friend, classmate, or family member.\\
5. Harry's Familiarity with the speaker, ranging from 0 (stranger) to 10 (a close friend).\\
6. Harry's Affection toward the speaker, ranging from $-10$ to 10.\\
Here is an example: \ldots\\
1. Before generating the response, read the above information and dialogue content carefully.\\
2. Do not generate a response that contradicts Harry Potter's attributes or his relations with the speaker.\\
3. Not every component of the background information may be useful; select the information needed to generate a concise and comprehensive response that matches Harry Potter's behavior in the dialogue.\\
4. When relation, familiarity, or affection information is unavailable, predict Harry's response from the dialogue context alone.
\end{promptbox}

\begin{promptbox}[teal]{RoleLLM Prompt}
\footnotesize
You are \{role\_name\}, your description is: \{role\_description\}. Now please answer some questions to accurately show your personality traits!
Your speaking style should fully imitate the personality role assigned to you!
Do NOT reveal you are an artificial intelligence model or a language model —
you must always remember you only have one assigned personality role.
Don’t be verbose, formal, or overly polite.
\end{promptbox}

\section{On the Unsuitability of Lexical Overlap Metrics for Role-Playing}
\label{sec:rouge}

As stated in the main text, we opted for a GPT-based pairwise evaluation for the \textbf{role-playing} task because standard lexical overlap metrics like ROUGE are unreliable for assessing performance in this domain \cite{lu-etal-2025-rolemrc}. To illustrate this, we calculated the ROUGE-L scores between the model-generated responses and the reference answers from the RoleLLM benchmark, using only the standard instruction-tuned models. Table~\ref{tab:rouge} presents these scores alongside the win rates from our GPT-4o-based evaluation.

The data reveals a significant mismatch between ROUGE-L scores and models’ role-playing performance. For instance, for LLaMA~3.1-8B and Qwen3-4B, using the Full System Prompt leads to a substantial increase in win rate but a \textit{decrease} in the ROUGE-L score. This discrepancy stems from the divergent and creative nature of role-playing, where a high-quality response can be lexically distant from the reference while still faithfully embodying the target persona. While ROUGE-L may have been serviceable when earlier models struggled to produce relevant content, the advancement of LLM agents necessitates more nuanced, fine-grained evaluation methods that prioritize semantic and stylistic fidelity.

To further validate our GPT-4o-based evaluation protocol, we conducted a parallel evaluation using a powerful open-source model, \texttt{Qwen3-30B-A3B-Instruct-2507}, as an alternative judge. The comparative results are presented in Table~\ref{tab:qwen}.

The results demonstrate a strong consistency in the relative performance ranking of the prompting methods across both judges. While Qwen3-30B tends to assign slightly higher scores, particularly in the 60-70\% win rate range, this may reflect a mild bias toward open-source models, noting that the reference answers were generated by GPT-4. Crucially, this slight scoring bias does not change the relative ordering of methods. The high agreement between the two judges supports the use of a strong LLM judge for this task, a practice supported by prior work showing that GPT-4's judgments align with human preferences over 80\% of the time~\cite{zheng2023judgingllmasajudgemtbenchchatbot,liu-etal-2023-g,huang-etal-2025-empirical}.

Finally, Table~\ref{tab:qwen} also includes the results for the ``Few-shot Examples + Prompt'' setting mentioned in Section~\ref{sec:Results}. As shown, incorporating few-shot examples into the full system prompt did not yield improvements and, in some cases, degraded performance. This observation motivates restricting few-shot experiments to the HPD and GSM8K datasets.

\begin{acks}
This research was partially supported by National Natural Science Foundation of China (No.62502404), Hong Kong Research Grants Council (Research Impact Fund No.R1015-23, Collaborative Research Fund No.C1043-24GF, RGC Research Fellow Scheme No.RFS2627-1S03, General Research Fund No.11218325, No.11212926), Institute of Digital Medicine of City University of Hong Kong (No.9229503), Huawei (Huawei Innovation Research Program), Tencent (CCF-Tencent Open Fund, Tencent Rhino-Bird Focused Research Program, Tencent University Cooperation Project), Kuaishou (CCF-Kuaishou Large Model Explorer Fund No.2025008, Kuaishou University Cooperation Project), Alibaba (CCF-Taobao \& Tmall Group TECH Kangaroo Fund 2026001), Didi (CCF-Didi Gaia Scholars Research Fund), and Bytedance.
\end{acks}
\section*{GenAI Usage Disclosure}

Generative AI tools were used for limited language polishing of author-written text, with all suggested edits reviewed and revised by the authors. The proposed framework, experimental design, and analysis are the original contributions of the authors.




\bibliographystyle{ACM-Reference-Format}
\balance
\bibliography{custom}

\end{document}